\documentclass[10pt,twocolumn,letterpaper]{article}
\usepackage{iccv}
\usepackage[utf8]{inputenc}
\usepackage[singlelinecheck=false]{caption}
\usepackage{times}
\usepackage{epsfig}
\usepackage{graphicx}
\usepackage{amsmath}
\usepackage{amssymb}
\usepackage{booktabs}
\usepackage[dvipsnames]{xcolor}
\usepackage[colorlinks = true, urlcolor  = Magenta]{hyperref}
\usepackage{array}
\usepackage{booktabs}
\newcommand{\prob}{VRC}
{}
{}
\usepackage{tabularx}
\usepackage{bm}
\usepackage{array}
\DeclareUnicodeCharacter{FB01}{fi}
\usepackage{multirow}

\def\iccvPaperID{10220} % *** Enter the ICCV Paper ID here

\iccvfinalcopy % *** Uncomment this line for the final submission

\ificcvfinal\fi

\begin{document}
\title{Few-Shot Visual Relationship Co-Localization}

\author{Revant Teotia*,~~~~Vaibhav Mishra*,~~~~Mayank Maheshwari*,~~~~Anand Mishra\\
Indian Institute of Technology Jodhpur \\
\textit{\{trevant,mishra.4,maheshwari.2,mishra\}@iitj.ac.in}\\
$*$: Contributed equally to the paper\\
\href{https://vl2g.github.io/projects/vrc/}{\textbf{https://vl2g.github.io/projects/vrc/}}
}
\maketitle
\ificcvfinal\thispagestyle{empty}\fi
\graphicspath{ {./figures/} }
\begin{abstract}
In this paper, given a small bag of images, each containing a common but latent predicate, we are interested in localizing visual subject-object pairs connected via the common predicate in each of the images. We refer to this novel problem as visual relationship co-localization or \prob{} as an abbreviation. 
\prob{} is a challenging task, even more so than the well-studied object co-localization task. This becomes further challenging when using just a few images, the model has to learn to co-localize visual subject-object pairs connected via unseen predicates. To solve \prob{}, we propose an optimization framework to select a common visual relationship in each image of the bag. The goal of the optimization framework is to find the optimal solution by learning visual relationship similarity across images in a few-shot setting. To obtain robust visual relationship representation, we utilize a simple yet effective technique that learns relationship embedding as a translation vector from visual subject to visual object in a shared space. Further, to learn visual relationship similarity, we utilize a proven meta-learning technique commonly used for few-shot classification tasks. Finally, to tackle the combinatorial complexity challenge arising from an exponential number of feasible solutions, we use a greedy approximation inference algorithm that selects approximately the best solution.

We extensively evaluate our proposed framework on variations of bag sizes obtained from two challenging public datasets, namely VrR-VG and VG-150, and achieve impressive visual co-localization performance.
\end{abstract}

\section{Introduction}
\label{sec:intro}
\begin{figure}
    \centering
    \includegraphics[scale=0.27]{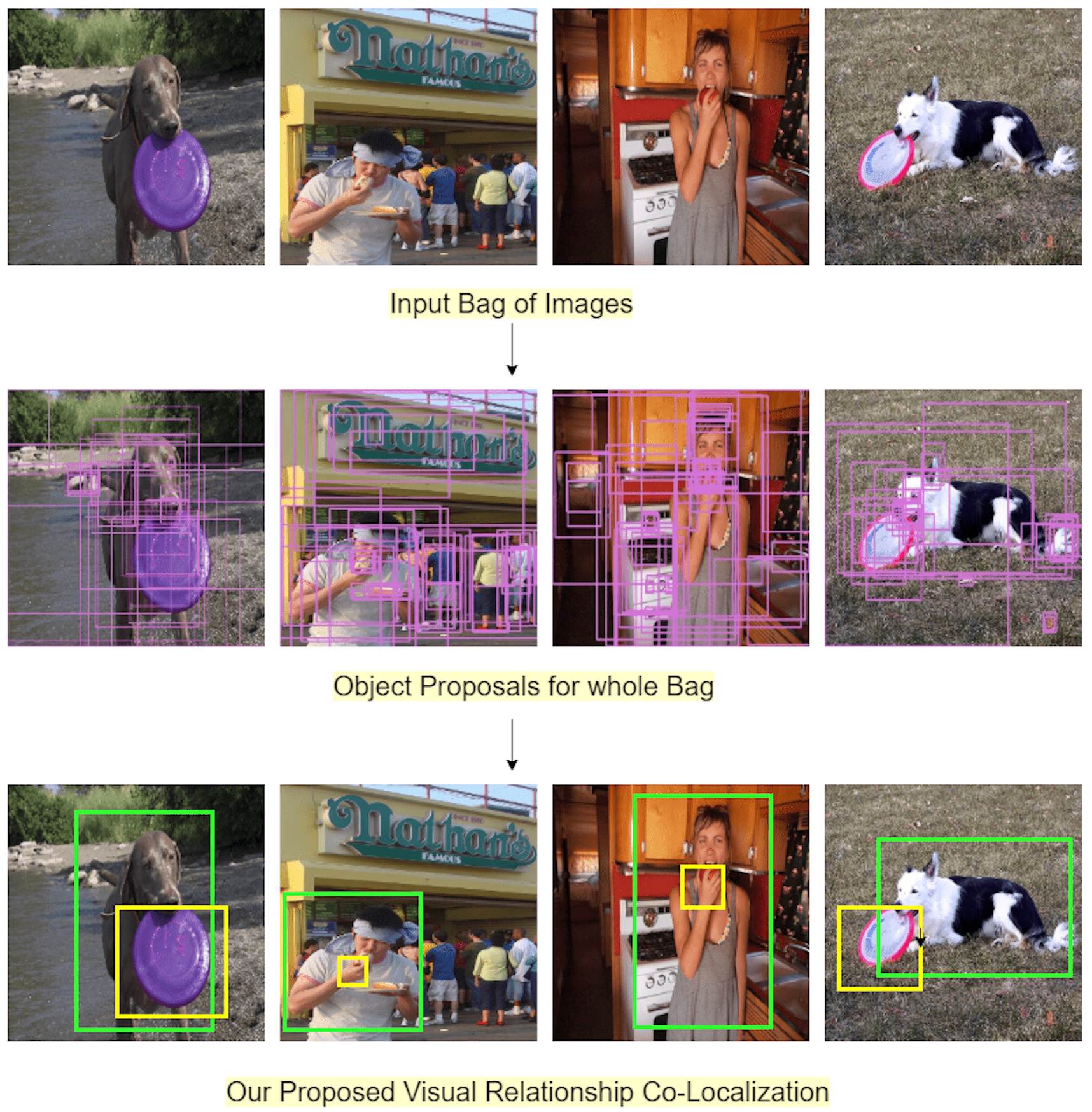}
    \caption{\label{fig:goal} \emph{Given a bag of four images as shown in the first row, can you find the visual subjects and objects connected via a common predicate?} Our proposed model in this paper automatically does that. In this illustration, the ``biting" predicate is present in all four images in the first row. Our proposed model localizes those visual subjects and objects in each image that are connected via ``biting" as shown in the third row. Note that the category name ``biting" is not provided to our approach. Here, green and yellow bounding boxes indicate the localized visual subject and objects respectively using our approach.\textbf{[Best viewed in color].} 
    }

\end{figure}

Localizing visual relationship $(<$subject, predicate, object$>)$ in images is a core task towards holistic scene interpretation~\cite{KrishnaCBF18, zhang2017visual}. Often the success of such localization tasks heavily relies on the availability of large-scale annotated datasets. \emph{Can we localize visual relationships in images by looking into just a few examples?} In this paper, towards addressing this problem, we introduce an important and unexplored task of \textbf{V}isual \textbf{R}elationship \textbf{C}o-localization (or \prob{} in short). \prob{} has the following problem setting: given a bag of $b$ images, each containing a common latent predicate, our goal is to automatically localize those visual subject-object pairs that are connected via the common predicate in each of the $b$ images. Note that, during both the training and testing phases, the only assumption is that each image in a bag contains a common predicate. However, its category, e.g. biting, is latent.

Consider Figure~\ref{fig:goal} to better understand our goal. Given a bag of four images, each containing a latent common predicate, e.g. “biting” in this illustration, we aim to localize visual subject-object pairs, such as (dog, frisbee), (man, hot dog), and so on, with respect to the common predicate in each of the images. VRC is significantly more challenging than well-explored object co-localization~\cite{silco,shaban2019learning,tang2014co} due to the following: (i) Common objects often share a similar visual appearance. However, common relationships can visually be very different, for example, visual relationships such as “dog biting frisbee” and “man biting hot dog” are very different in visual space. (ii) Relationship co-localization requires both visual as well as semantic interpretation of the scene. Further, VRC is also distinctly different from visual relationship detection (VRD) that aims to estimate the maximum likelihood for $(<$subject, predicate, object$>)$ tuples from a predeﬁned ﬁxed set of visual relationships common across the train and test sets. It should be noted that test predicates are not provided even during the training phase of \prob{}. Therefore, the model addressing \prob{} has to interpret the semantics of unseen visual relationships during test time.

Visual relationship co-localization (\prob{}) has many potential applications, examples include automatic image annotation, bringing interpretability in image search engines, visual relationship discovery. In this work, we pose \prob{} as a labeling problem. To this end, every possible visual subject-object pair in each image is a potential label for common visual subject-object pair. To get the optimal labeling, we define an objective function parametrized by model parameters whose minima corresponds to visual subject-object pairs that are connected via a common latent predicate in all the images. To generalize well on unseen predicates, we follow the meta-learning paradigm to train the model. Just as a good meta-learning model learns on various learning tasks, we train our model on a variety of bags having different common latent predicates in each of them so that the model generalizes to new bags. We use a greedy approximation algorithm during inference that breaks down the problem into small sub-problems and combines the solutions of sub-problems greedily.

To evaluate the performance of the proposed model for \prob{}, we use two public datasets, namely VrR-VG~\cite{liang2019vrr} and VG-150~\cite{xu2017scene}.
Our method achieves impressive performance for this challenging task. This is attributed to our principled formulation of the problem by defining a suitable objective function and our meta-learning-based approach to optimize it.  Further, we present several ablation studies to validate the effectiveness of different components of our proposed framework. On bag size $=4$, we achieve 76.12\% co-localization accuracy on unseen predicates of VrR-VG~\cite{liang2019vrr} dataset.

The contributions of this paper are two folds: (i) We introduce a novel task -- \prob{} (Visual Relationship Co-Localization). \prob{} has several potential applications and is an important step towards holistic scene interpretation. (ii) Inspired by the recent success of the meta-learning paradigm in solving few-shot learning tasks, we propose a novel framework for performing few-shot visual relationship co-localization. Our framework learns robust representation for latent visual predicates and is efficacious in performing visual relationship co-localization with only a few examples. %The implementation of our proposed approach is available at: 

\section{Related Work}
\label{sec:relWork}
\noindent\textbf{Object Co-localization:}
Object localization~\cite{chen2017discover, jie2017deep,shen2018generative,zhuang2017attend} is an important and open problem in computer vision. To localize object overlap between two or more images, object co-localization has been introduced. In an early work, Tang et al.~\cite{tang2014co} have proposed the box and image model in an optimization framework to address object co-localization. In their formulation, both the models complement each other by helping in selecting clean images and the boxes that contain the common object. Towards addressing the limited annotated data issue, the recent works~\cite{silco,shaban2019learning} have opted for the lane of few-shot learning. Hu et al.~\cite{silco} localize a common object across support and a query branch. Whereas Shaban et al.~\cite{shaban2019learning} form bags of images, and then find common objects across all the images in a bag. While object co-localization is an interesting task, visual relationship co-localization requires a visual as well as a semantic understanding of the scene. To the best of our knowledge, few-shot visual relationship co-localization has not been studied in the literature.  
\newline
\newline
\noindent\textbf{Visual Relationship Detection (VRD):}
It is an instrumental task in computer vision due to its utility in comprehensive scene understanding. To get the predicted relationship label in the image, Zhang et al.~\cite{zhang2017visual} used the spatial, visual, and semantic features. This approach is limited to detecting those relationships that are available during the training and does not generalise on unseen relationships. Another method~\cite{zhang2019large} project the objects and relations into two different higher dimensional spaces and ensures their semantic similarity and distinctive affinity by using multiple losses. Zhang et al.~\cite{zhang2019graphical} introduced a new graphical loss to improve the visual relationship detection. Zellers et al.~\cite{zellers2018neural} used a network of stacked bidirectional LSTMs and convolutional layers to parse a scene graph and, in between, detect various relationships in the image. Many recent approaches have also benefited from the advancements in graph neural networks~\cite{li2018factorizable}. As compared to visual relationship detection, we are distinctively different, as discussed in the introduction of this paper. 
\newline
\newline
\noindent\textbf{Meta Learning for Few-Shot Learning:} 
Few-shot learning methods~\cite{andrews2002support,carbonneau2018multiple,finn2017modelagnostic,wang2020generalizing} are being studied and explored significantly for both computer vision~\cite{doran2014theoretical,pathak2014fully} and natural language processing~\cite{BrownMRSKDNSSAA20,han2018fewrel,ravi2016optimization,garcia2018fewshot,wang2020generalizing,yan2018few}. There are two major groups of methods towards solving the few-shot learning problem: (i) metric-based and (ii) model-based methods. Siamese Networks~\cite{koch2015siamese} which uses a shared CNN architecture for learning the embedding function and weighted L1 distance for few-shot image classification, Matching Network~\cite{vinyals2016matching} which uses CNN followed by an LSTM for learning the embedding function, Prototypical Network~\cite{snell2017prototypical} which uses CNN architecture with a squared L2 distance function and Relation Network~\cite{RelationNet} which proposed to replace the hand-crafted distance metrics with a deep distance metric to compare a small number of images within episodes, are examples of metric-based approaches. Model-based approaches generally depend on their model design. MetaNet \cite{munkhdalai17a} is an example of a model-based few-shot learning approach that enables rapid generalization by learning meta-level knowledge across multiple tasks and shifting its inductive biases via fast parameterization. We use a metric-based approach viz. Relations Network for learning similarity between visual relationship embeddings in our optimization framework. 

% Gradient-based meta-learners have two models, namely base-learner and meta-learner, the meta-learner learns across episodes and the base-learner, which is learned inside the episode, the aim is to learn the optimization of the model weights. \textcolor{red}{We adopt a metric-based meta-learning-based approach in this work to perform a few-shot visual relationship co-localization.}  

\begin{figure*}[!t]
    \centering
    \includegraphics[width=17cm]{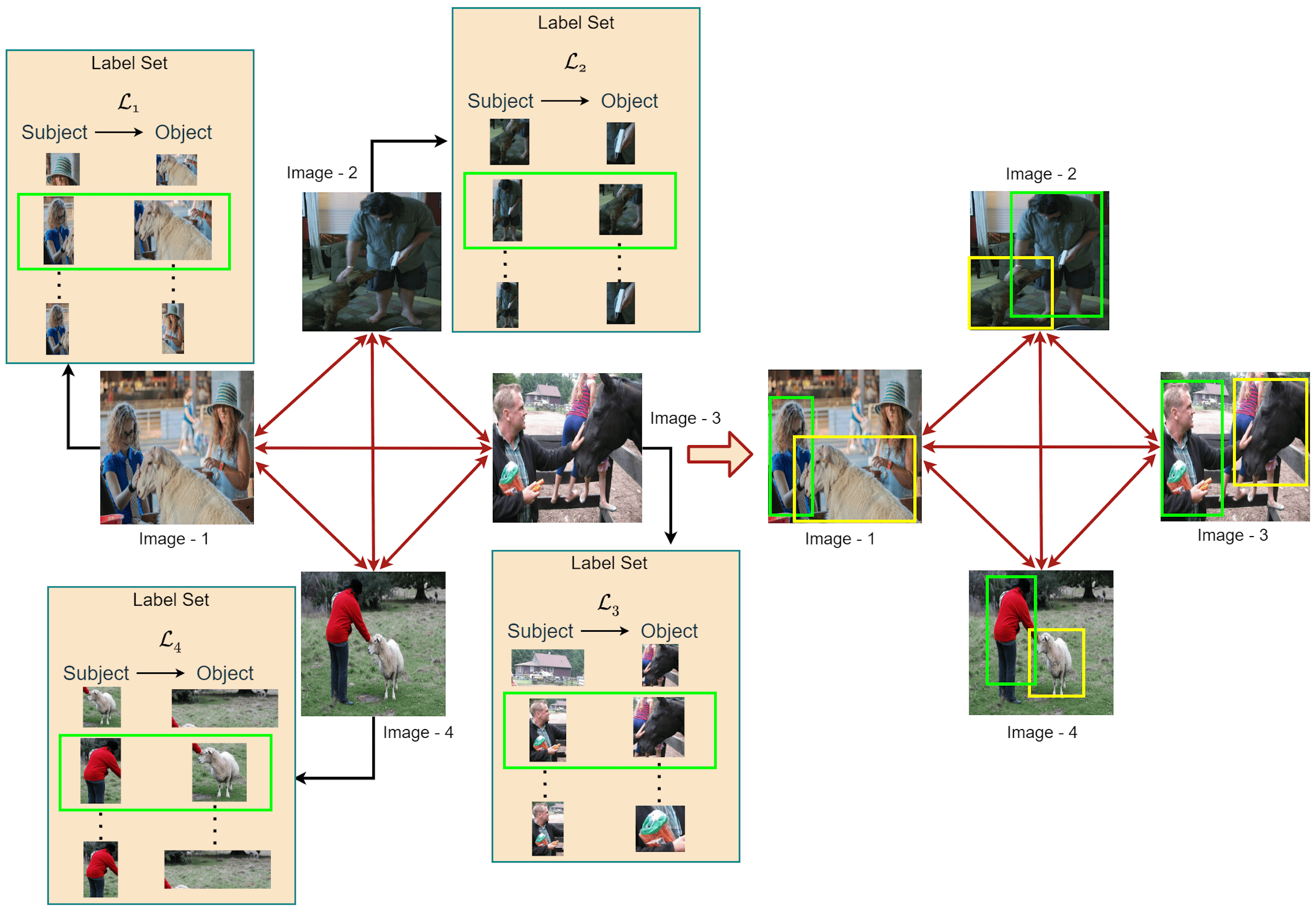}
    \caption{ \label{fig:graph_min}\textbf{\prob{} as a labeling problem.} Given a bag of $b$ images ($b = 4$ in this illustration), we construct a fully connected graph by denoting each image as a vertex. All the pairs of object proposals in each image constructs the label set for each vertex. The goal is to find a labeling such that the labels representing common latent predicate are selected for each image, e.g., ``petting" in this illustration. We solve this problem by minimizing a corresponding objective function. Refer to Section~\ref{sec:problem} for more details. \textbf{[Best viewed in color].} }
\end{figure*}
\begin{table}[!t]
    \centering
    \renewcommand*{\arraystretch}{1.2}
    \begin{tabular}{c|l}
        Notation & Meaning \\
        \toprule
         ${\cal L}_u$& Label set for Image-$u$\\
         $l_{so} \in {\cal L}_u$ & Visual relationship \\
         $b$& Bag size\\
         $p_u$& Number of object proposals in Image-$u$\\
         $B_i^u$& $i$th object proposal in Image-$u$\\
         $P_{so}$&     \vtop{\hbox{\strut Latent predicate connecting proposals}\hbox{\strut $B_{s}^u$ and $B_{o}^u$}}\\
         $P_{so}^*$& Common latent predicate \\
         $f_\phi(\cdot)$& Relationship embedding network \\
         ${R_{\bm{\theta}}}(\cdot,\cdot)$ & Visual relationship similarity\\
         $\bm{\theta}$ & Model parameters\\
         \bottomrule
    \end{tabular}
    \caption{\label{tab:not}Notations used in this paper.}
\end{table}
\section{Approach}
\label{sec:problem}
Given a bag of $b$ images, $\{I_u\}_{u=1}^{b}$ such that each image of the bag $I_u$ contains a latent common predicate that is present across all the images in the bag, our goal is to find the set $O$ such that $O = \{(B^{u}_i, B^{u}_j)\}_{u=1}^{b}$ where each tuple $<B^{u}_i, B^{u}_j>$ corresponds to object proposal pairs in $u$th image that are connected via the common predicate in the bag. Here, $B^{u}_i$ and $B^{u}_j$ are the bounding boxes over visual subject and object respectively. Table~\ref{tab:not} shows the meaning of major notations used in this paper.

\subsection{\prob{} as a Labeling Problem}
\label{sec:graph_labeling}
We pose \prob{} as a labeling problem. To this end, given a bag containing $b$ images, we construct a fully connected graph $G = \{V,E\}$ where $V = \{I_u\}_{u=1}^{b}$ is a set of vertices such that each vertex corresponds to an image. The potential label set for each vertex is a set of all possible pairs of object proposals\footnote{Object proposals should not be confused with the object in a visual relationship.} obtained from the corresponding image. 
Given this graph and label sets, the goal is to assign one label to each vertex of the graph (or equivalently to each image in the bag) such that visual subject-object pair connected via the latent common predicate $P_{so}^*$ is assigned to each image.

The labeling problem formulation for the visual relationship co-localization using an illustrative example is shown in Figure~\ref{fig:graph_min}. Here, we show four images in a bag, i.e., bag size $b = 4$. Each image is represented as a vertex in a fully-connected graph $G$. To obtain a label set for each of these vertices (or equivalently each image), we first obtain object proposals using Faster R-CNN~\cite{FasterRCNN2015}. 
Let $B = \{ B_{i}^u\}_{i=1}^{p_u}$ be a set of object proposals obtained for Image-$u$, for example, in Figure~\ref{fig:graph_min}, we get \emph{bounding boxes} for ``woman", ``sheep", ``hat", ``bucket", etc. as object proposals for Image-1. Here $p_u$ is the number of object proposals in Image-$u$.
Given these, the label set of this vertex will contain all possible ordered pairs of object proposals. In other words, the cardinality of this label set is equal to $p_u \times (p_u-1)$. 

Further, each ordered pair of the object proposals is connected via a latent predicate. Examples of latent predicate in Image-1 (ref. Figure~\ref{fig:graph_min}) are petting, wearing, etc. These predicates define visual relationships such as ``$<$woman, petting, sheep$>$", ``$<$woman, wearing, hat$>$", etc. Suppose  $<B_s^u, P_{so}, B_o^u>$ represents that object proposals $B_s^u$ and $B_o^u$ of image-$u$ that are connected via a hidden predicate $P_{so}$. Then, the label set for Image-$u$ or equivalently corresponding vertex-$u$ is given by:
\begin{equation}
\begin{split}
{\cal L}_u = \{<B_s^u, P_{so}, B_o^u>~|~s \neq o~\text{and}~(B_s, B_o)~\text{is an} \\ \text{ordered pair of object proposals in image-}u ~\text{and}~\\P_{so}~\text{is a latent predicate.}\}
    \end{split}
\end{equation}

A label $l_{u(s,o)} =~<B_s^u, P_{so}, B_o^u>~\in~ {\cal L}_u$ is an instance (or member) of label set for vertex-$u$. For simplifying the notation, we write $l_{u(s,o)}$ as $l_{ut}$ from here onwards  where $t$ varies from $1$ to $|{\cal L}_u|$. Further, the optimal label, i.e., the visual subject-object pair that are connected via a ``common" latent predicate ${P^*_{so}}$ in image-$u$ is represented by: $l^*_{ut}$. 
In Figure~\ref{fig:graph_min}, $P^*_{so}$ = ``petting"  with visual relationship tuples $<$woman, petting, sheep$>$, $<$man, petting, dog$>$, $<$man, petting, horse$>$, $<$man, petting, sheep$>$ in Image-1 to 4 respectively.  
Recall that the goal of the labeling problem is to assign the optimal labels to all of the bag images or, in other words finding an optimal pair of subject and object bounding boxes $<{B^u_s}^*, {B^u_o}^*>$ for each bag image.

\noindent\textbf{Formulation for the optimal labeling}:
To solve the labeling problem, we define the following objective function whose minima corresponds to optimal labeling for \prob{}, i.e., localizing the visual subject-object pairs in each image of a bag that are connected via the common latent predicate:

\begin{equation}
  \Psi = \sum_{u=1}^{b} \Big( \min_t \Psi_u (l_{ut}) + \sum_{v=1}^{b, u \neq v}  \min_{t_1, t_2} \Psi_{uv} (l_{ut_1}, l_{vt_2}, \bm{\theta}) \Big).
\label{eq:obj}
\end{equation}

In this objective function there are two terms: (i) Unary term $\Psi_u (l_{ut})$ which represents cost of assigning a label $l_{ut} = <B_s^u, P_{so}, B_o^u>$ to image $u$. Since given an image, any subject-object pair is considered to be equally likely. Therefore, this term of the objective function does not contribute to the optimization.\footnote{We write a ‘general
form of the cost function (unary+pairwise)’ to emphasize
that ‘theoretically’ the likelihood of a subject-object pair in
an image could also contribute to the optimization.} (ii) Pairwise term $\Psi_{uv} (l_{ut_1}, l_{vt_2}, \bm{\theta})$ represents the cost of image-$u$ taking a label  $l_{ut_1} =~<B_s^u, P_{so}, B_o^u>$ and image-$v$ taking a label $l_{vt_2} =~<B_s^v, P_{so}, B_o^v>$. Here $\bm{\theta}$ is a learnable model parameter that needs be learnt from few examples. We use a neural model to learn these parameters. We describe this neural model in Section~\ref{sec:meta}. Further, the pairwise term of optimization should be defined in such a way that it is lower when hidden predicates $P_{so}$ of $l_{ut_1}$ and $l_{vt_2}$ are semantically similar, and higher otherwise. We compute this pairwise term in Equation~\ref{eq:pair_cost}. Further, to compute this pairwise term, we need to first learn a robust semantic encoding of $l_{so}$ given pair of object proposals $B_s$ and $B_o$ which are represented using concatenation of bounding box coordinates, Faster-RCNN fc6 features, and object class scores. In other words, we wish to learn visual relation embedding as follows:
\begin{equation}
    f_{l_{so}} = f_{\Phi}(B_s, B_o),
    \label{eq:vre}
\end{equation}
where $f_{\Phi}$ denotes our visual relationship embedding network parameterized by $\Phi$ and $f_{l_{so}}$ is the encoding of visual relationship $l_{so}$. We use a popular relationship encoding network viz. VTransE~\cite{zhang2017visual} for computing relationship embedding.

\subsection{Learning to Label with Few Examples}
\label{sec:meta}
In our problem setting, to be able to generalize well on new bags, the model should be able to learn similarity between visual relationships even when looking into small-size bags at a time. This is usually referred to as a few-shot setting. Many learning paradigms exist for addressing the problem in this setting. We choose Meta-Learning~\cite{hospedales2020meta, santoro2016meta} which is one of the most successful approaches. To be specific, we use one of the metric-based meta-learning approaches viz. Relation Network~\cite{RelationNet} to learn the similarity between visual relationships as follows.

Given a pair of visual relationships $l_i$ and $l_j$, we first obtain their representations $f_{l_{i}}$ and $f_{l_{j}}$ respectively using the Equation~\ref{eq:vre}. Then we calculate similarity score between these representations as follows: %$(R_{\bm{\theta}}(f_{l_i}, f_{l_j}))$ we use the following equation %\ref{eq:pair}.   

\begin{equation}
% \label{eq:relascore}
R_{\bm{\theta}}(f_{l_i}, f_{l_j}) = w^T{K(f_{l_i}, f_{l_j})}+b,
\label{eq:pair}
\end{equation}
where $w$ is a learnable weights matrix and $b$ is the bias vector. Further, $K$ is computed as follows:
\begin{multline}
\label{eq:K_calculation}
   K(f_{l_i}, f_{l_j}) = \tanh(W_1([f_{l_i}; f_{l_j}])+b_1)\\\sigma(W_2[f_{l_i}; f_{l_j}]+b_2)
+ ((f_{l_i}+f_{l_j})/2),
\end{multline}
where $W_1, W_2$ are two learnable weight matrices, $b_1, b_2$ represent the bias vectors. Further, $\tanh$ and $\sigma$ represent the hyperbolic tanh and sigmoid activation function respectively. Here, instead of only using the mean of visual relationship features, we also add a widely used learnable gated activation~\cite{ramachandran2017swish,OordKEKVG16} to get a better feature combination.

We train the Relation Network parameters using episodic binary logistic regression loss. To this end, for each bag, we create all possible pairs of $l_i$ and $l_j$ such that they belong to different images in the bag. A pair of $l_i$ and $l_j$ is positive if the predicates of both are the same as the common latent predicate of the bag; otherwise, it is negative. We finally compute the pairwise cost as negative of the learned similarity metric, i.e., 
\begin{equation}
% \label{eq:relascore}
\Psi_{uv} (l_{ut_1}, l_{vt_2}, \bm{\theta}) = -R_{\bm{\theta}}(f_{l_{ut_1}}, f_{l_{vt_2}}).
\label{eq:pair_cost}
\end{equation}

\subsection{Inference}
\label{sec:inference}
The problem of finding the global optimal solution for the optimization function in Equation~\ref{eq:obj} is an NP-hard problem. The cardinality of the label set of an image is $p_u \times (p_u-1)$ where $p_u$ is number of object proposals in image-$u$. Therefore, a brute force technique to find the optimal solution to this labeling problem will take $O\big(\prod_{u=1}^{b} p_u^2\big)$ time.  We adopt a greedy inference algorithm proposed by Shaban et al.~\cite{shaban2019learning} due to its proven superiority over other approximation algorithms available for solving these kinds of problems~\cite{bergtholdt2010study,kolmogorov2006convergent}.

\begin{table*}[!t]
\renewcommand{\arraystretch}{1.2}
\centering
\begin{tabular}{l|c|c|c|c|c}
\toprule
\multirow{2}{*}{Variations of our approach}
& \multirow{2}{*}{Bag size} & \multicolumn{2}{c|}{VrR-VG} & \multicolumn{2}{c}{VG-150} \\ \cline{3-6} 
                                              &                           & Bag-CorLoc  (\%)       &     VR-CorLoc  (\%)     & Bag-CorLoc (\%)     & VR-CorLoc  (\%)       \\ \hline
\multirow{3}{*}{Concat + Cosine Similarity}
& 2
& 55.90 & 72.16 & 50.00  & 71.42
\\ \cline{2-6} 
& 4                         
& 31.57 & 70.86 & 24.40  & 65.58
\\ \cline{2-6} 
& 8
& 30.65  & 76.85  & 18.75  & 67.33
\\ \hline
\multirow{3}{*}{VTransE + Cosine Similarity}    
& 2               
& 59.84 & 73.34  & 55.67  & 74.90  
\\ \cline{2-6} 
& 4 
& 36.23  & 74.20  & 33.45  & 71.78
\\ \cline{2-6} 
& 8 
& 34.64  & 82.56  & 26.67  & 70.85
\\ \hline
\multirow{3}{*}{Concat + Relation Network}        
& 2                         
& 61.72  & 75.61  & 54.55 & 71.85
\\ \cline{2-6} 
& 4                         
& 35.28 & 74.02 & 38.62  & 72.19 
\\ \cline{2-6} 
& 8                        
& 31.24 & 76.38 & 29.15 & 75.55 
\\ \hline
\multirow{3}{*}{\textbf{Our best model}}   
& 2                         
& \textbf{63.40} & \textbf{78.99} &  \textbf{61.10} & \textbf{75.82}
\\ \cline{2-6} 
& 4 
& \textbf{48.06} & \textbf{76.12} &  \textbf{42.30} & \textbf{79.15} 
\\ \cline{2-6} 
& 8                         
&  \textbf{45.48}  & \textbf{84.07} & \textbf{37.61}  & \textbf{79.96}        
\\ \bottomrule
\end{tabular}
\caption{\label{tab:main}\textbf{Visual Relationship Co-localization results on unseen predicates.} We observe that our best model which uses VTransE for representing visual relationships and relation network for computing relationship similarity outperforms other variants by a significant margin. Impressive visual relationship co-localization performance by our approach verifies the effectiveness of relationship embedding and metric-based meta-learning approach to compute visual relationship similarity as components in our approach and our overall optimization framework. Note: We sampled three different sets of training bags to evaluate our model and found that VR-CorLoc only varied by the standard deviation of $\pm2.7\%$. }
\end{table*}

\section{Experiments and Results}
\label{sec:expts}
\subsection{Datasets and Experimental Setup}
\label{sec:data}
To quantitatively study the robustness of our proposed approach, we have used the following two public datasets for all our experiments.

    \noindent \textbf{(i) VrR-VG}~\cite{liang2019vrr}: Visually relevant relationships dataset (VrR-VG in short) is derived from the Visual Genome~\cite{VisualGenome2016} by removing all the statistically and positionally-biased visual relationships. It contains 58,983 images, 23,375 visual relationship tuples, and 117 unique predicates. Out of these 117 predicates, we use randomly chosen 100 predicates for training and the remaining 17 predicates for testing. 
    
    \noindent \textbf{(ii) VG-150}~\cite{xu2017scene}: To test the robustness of our approach, we further show results on VG-150. This dataset contains 150 object categories and 50 predicate classes. Out of the 50 predicates, we use 40 and 10 for training and testing, respectively.

To obtain object proposals for an image, we use Faster R-CNN~\cite{FasterRCNN2015} trained on Visual Genome~\cite{VisualGenome2016}. We then select the top-100 most confident object proposals after performing a non-max suppression with a 0.5 intersection over union (IoU) threshold. To create the label set for an image, we consider all possible ordered pairs of object proposals for that image as candidates for the common visual relationship. Since we consider top-100 object proposals per image, we get 9900 $(= 100 \times (100-1))$  candidates for visual subject-object pairs in each image. 

Further, we train VTransE~\cite{zhang2017visual} using training predicates to obtain visual relationship embeddings. To create an image bag of size $b$, we first select a predicate and then pick $b$ images from the dataset such that each of the $b$ images has at least one visual relationship with the selected predicate. In this way, we get a bag in which all the images share a common predicate. We create 10,000 training bags and 500 testing bags using disjoint set of training and testing predicates respectively.

\noindent\textbf{Performance Metrics:}
Following the widely-used localization metric CorLoc~\cite{deselaers2010localizing}, we use the following two performance metrics to evaluate the performance of our approach:

\noindent\textbf{(i) Visual Relation-CorLoc:} In an image, a visual relationship candidate prediction is considered to be correct if both its visual subject and visual object localization are correct.\footnote{An object proposal is considered to be correct if it has greater than 0.5 $IoU$ with the target ground-truth bounding box.} \emph{VR-CorLoc is defined as the fraction of test images for which visual subject-object pairs are correctly localized.}
    
\noindent\textbf{(ii) Bag-CorLoc:} If the common visual relations are correctly predicted for all of the bag images, then we consider that bag to be correctly predicted. \emph{Bag-CorLoc is defined as the fraction of the total number of bags for which the visual subject-object pairs are correctly localized for all of its images.}

% ######################################################################################################################
\begin{table*}[!t]
\centering
\begin{tabular}{l|c|c|c|c|c|c|c|c|c|c|c|c}
\toprule
\begin{tabular}[l]{@{}l@{}}
Variations of\\ our approach $\rightarrow$   
\end{tabular}
                                                                & \multicolumn{3}{c|}{\begin{tabular}[c]{@{}l@{}}Concat + Cosine\end{tabular}} & \multicolumn{3}{c|}{\begin{tabular}[c]{@{}l@{}}VtransE+ Cosine\end{tabular}} & \multicolumn{3}{c|}{\begin{tabular}[c]{@{}l@{}}Concat+ Rel. Net\end{tabular}} & \multicolumn{3}{c}{\begin{tabular}[c]{@{}l@{}}Our best model\end{tabular}} \\ \hline
Supervision $\downarrow$                                                                & \multicolumn{3}{c|}{Bag Size}                                                  & \multicolumn{3}{c|}{Bag Size}                                                  & \multicolumn{3}{c|}{Bag Size}                                                       & \multicolumn{3}{c}{Bag Size}                                                        \\ \cline{2-13}
                                                                           & 2                        & 4                        & 8                        & 2                        & 4                        & 8                        & 2                          & 4                          & 8                         & 2                          & 4                          & 8                          \\ \hline
No supervision                                            
&72.16 &70.86 & 76.85
&73.34 & 74.20 & 82.56                          
&75.61 & 74.02 & 76.38
&78.99 & 76.12 & 84.07                        \\ \hline
\begin{tabular}[l]{@{}l@{}}Subject Fixed \end{tabular} 
                        
&76.82 & 78.66 &  \textbf{81.27}
&80.37 &\textbf{83.12}&83.58  
& \textbf{81.07} & \textbf{82.88}  &\textbf{84.60}                   
&83.90 &\textbf{88.25} &86.67                            \\ \hline
\begin{tabular}[l]{@{}l@{}}Subject-Object\\ in one image\end{tabular}                         

& \textbf{77.03} & \textbf{80.20} &  79.42
&\textbf{83.33}&82.40& \textbf{84.07}
&79.29 &81.69&   81.45
&\textbf{87.44} &84.46&\textbf{86.95} 
\\ \bottomrule
\end{tabular}
\caption{\label{tab:ablation} \textbf{Effects of weak supervision on \prob{}.} We observe that just by giving a weak form of supervision, e.g., fixing subject in all images of bag or fixing subject and object in one image of the bag, the visual relationship co-localization performance (\% VR-CorLoc) increases significantly using our approach. Refer Section~\ref{sec:base} and Section~\ref{sec:resNDis} for more details.}
\end{table*}

\subsection{Ablations and Different Problem Settings}
\label{sec:base}
\noindent \prob{} being a novel task, we do  not have any direct competitive method to compare with our proposed approach. However, to justify the utility of different modules of our approach (also referred as our best model) and to show robustness on few-shot visual relationship localization, we perform the following ablation studies:

\noindent\textbf{(i) VtransE + cosine Similarity}: As the first ablation, to verify the utility of Relation Network that we use to compute the similarity between two of the relationship embeddings $f_{l_i}$ and $f_{l_j}$ , we replace the it by a cosine similarity.
    % \begin{equation}
    %     CosineSim(f_{l_i}, f_{l_j}) = \frac{f_{l_i} \cdot f_{l_j}}{||f_{l_i}|| \times ||f_{l_j}||}.
    % \end{equation}

\noindent\textbf{(ii) Concat Embedding + Relation Network}: To verify the utility of  relationship embedding encoder network in our best model viz. VTransE, we replace it with just a trivial concatenation of subject and object embeddings, i.e., $f_{l_i} = [s; o]$ where $s$ and $o$ represent the concatenation of Faster R-CNN features, bounding box coordinates, and object class probability scores of subject and object respectively. The rest of the method is identical to ours.

\noindent\textbf{(iii) Concat Embedding + cosine Similarity}: In this ablation, we replace both the vital components of our approach, i.e., VtransE and Relation Network, by cocat embedding and cosine similarity respectively.

Further, in the original problem setting of \prob{}, only a bag of images is provided (no supervision). While we perform the experiment in this challenging setting, we also relax the problem setting a bit as follows in conducting additional experiments:

\noindent\textbf{(i) Visual subjects in all the images are given:}
In this setting, along with the bag of images, we assume that a bounding box for the visual subject is also provided in each image. Our goal is to only co-localize those visual objects that connect the given subject via a common predicate in all the images of the bag.
    
\noindent\textbf{(ii) Both visual subject-object in one image is given:} In this setting, both visual subject and object bounding boxes corresponding to the common latent predicate are provided but only for one image of the bag. Given this, our goal is to co-localize visual subjects and objects in the remaining images of the bag.

We show results of these ablations and problem setting variations on datasets presented in Section~\ref{sec:data}, and compare them against our best model in the next section.

\subsection{Results and Discussion}
\label{sec:resNDis}
We first perform a quantitative analysis of our proposed approach in Table~\ref{tab:main}. We report Bag-CorLoc and VR-CorLoc (refer Section~\ref{sec:data}) in \% for bag size varying from 2 to 8. We observe that by the virtue of the right choice of visual relationship embedding technique and metric-based meta-learning approach in our principle optimization framework, our best model achieves 45.48\% Bag-CorLoc and 84.07\% VR-CorLoc on VrR-VG on bag size = 8. Such an impressive visual relationship co-localization verifies the efficacy of our proposed approach. 

Further, to justify our choice of VTransE for learning visual relationship embedding and Relation Network to compute the similarity between visual relationship embeddings, we perform ablations by replacing VTransE with a simple concatenation of subject and object features and Relation Network by cosine similarity. As shown in Table~\ref{tab:main}, our framework with a simple visual relation embedding such as concatantion of subject-object features and a simple similarity computation such as cosine similarity achieves reasonable performance. This can be attributed to our meta-learning-based optimization approach. The choice of VTransE and Relation Network modules in our framework (see our best model, last row) further improves the performance of visual relationship co-localization. We notice similar trend in visual relationship co-localization performance in VG-150 as well.

We also perform extensive experiments with minor tweaks in the original setting of \prob{} by relaxing it a bit. We have shown VR-CorLoc for all those experiments in Table~\ref{tab:ablation} on the VrR-VG dataset. We observe that once we relax the strictness in problem setting a little bit, in other words, by providing subject bounding boxes, the VR-CorLoc increases significantly for each of the ablation and prominently if we see our approach for bag size two and four, it increases to 83.90\% and 88.25\% from 78.99\% and 76.12\% respectively. In the other scenario where we relax the condition by only giving subject and object bounding boxes for only one image in the bag, the VR-CorLoc score increases to 87.44\%  and 84.46\% from 78.99\% and 76.12\% for bag size two and four, respectively. These results shows that by providing slightly more supervision (either annotating bounding boxes for subject corresponding to a common predicate in all the images or annotating subject-object pair corresponding to a common predicate in one image), the visual relationship co-localization of our approach significantly improves. 

\begin{figure*}[!t]
    \centering
    \includegraphics[width=17cm]{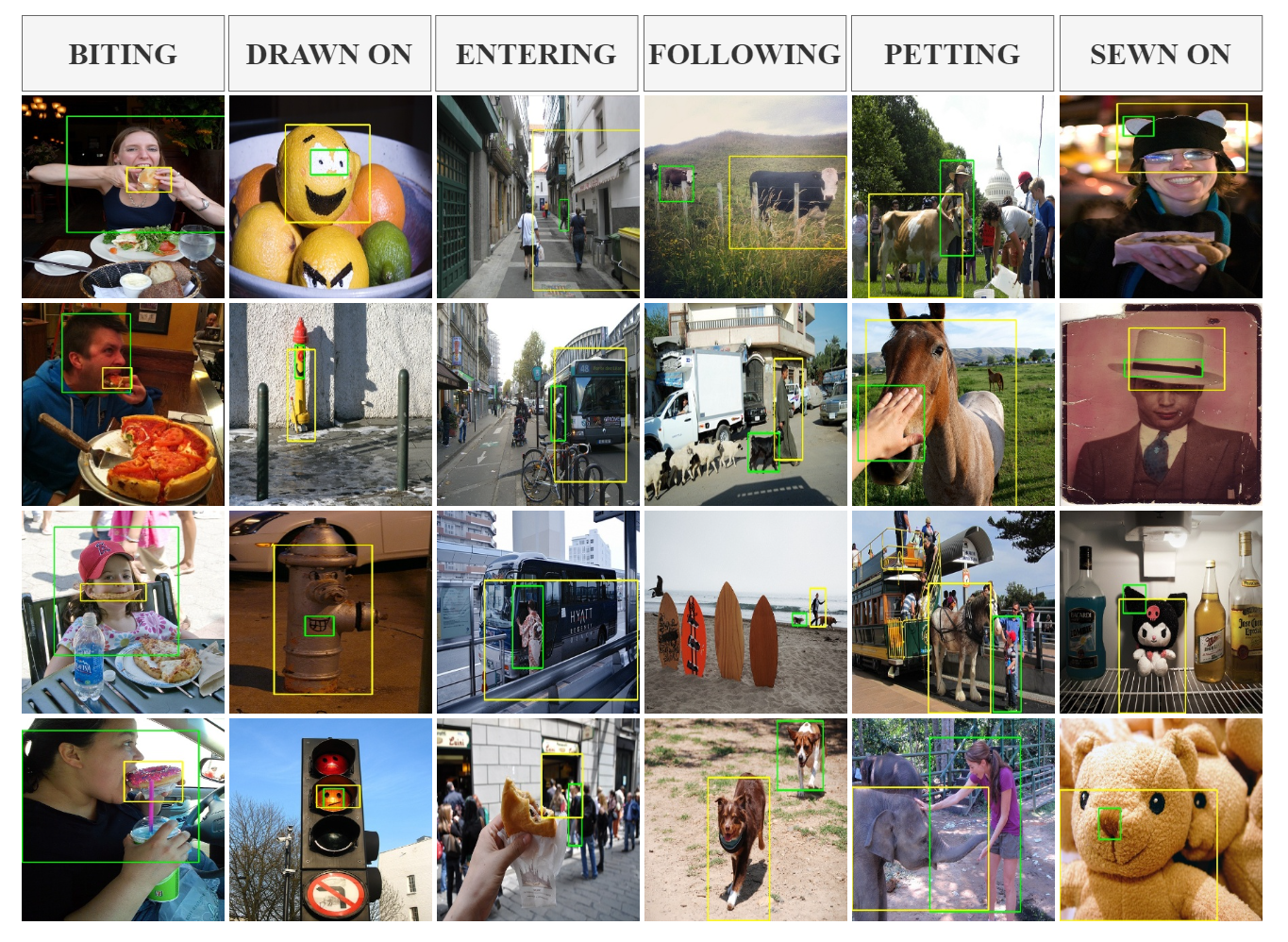}
    \caption{\label{fig:visRes} We show some of the qualitative results of our approach on the VrR-VG dataset.  Each column is a bag of images (bag size = 4), having a common latent predicate in all its images. The common latent predicate is written on top of each column. Our approach localizes the visual subject-object pairs in each image of the bag, which are connected through that common latent predicate by drawing bounding boxes around them. The green and yellow bounding boxes correspond to the localized visual subject and object, respectively. It should be noted that all of these predicates are never seen during the training phase. \textbf{[Best viewed in color and 200\% zoom in]}. }
\end{figure*}
A selection of visual relationship co-localization results by our approach is shown in Figure~\ref{fig:visRes}.\footnote{More visual results are presented in Supplementary Material.} 
Here we show a bag of images in each column. The subject and object co-localization on these bags is shown using bounding boxes of green and yellow colors, respectively. We observe that our approach successfully co-localizes the visual subject and objects connected via a latent predicate by just looking into four images in the bag. Specifically, consider the fourth column where the latent predicate is \emph{Following}. Our approach co-localizes subject and object following to each other, for example ``a cow \emph{following} to another cow" in row-1, ``a sheep \emph{following} to a man" in row-2 and so on.
Given that our model has not seen the predicate \emph{following} during the training and there are different combinations of subject and object following each other, these results are encouraging. Note that all the relationships shown in Figure~\ref{fig:visRes} are `unseen' during the training phase. 

As the first work towards visual relationship co-localization, we focus on co-localizing only one common visual relationship. Our primary dataset VrR-VG does not contain visually-trivial relationships, e.g. ‘car has wheels’, ‘man wearing shirt’, and as the bag size grows ($2 \rightarrow 4 \rightarrow 8$), it naturally becomes less likely to have more than one common predicate present in `all' of the images. For example in VrR-VG test set, only 68/500, 1/500, 0/500 bags of sizes 2, 4, 8 respectively have more than one common predicate. In cases, where there are more than one common predicates, for example in VG-150, our method predicts the one which corresponds to minimum pairwise cost and drops the other common predicates. This results in slightly inferior performance on dataset containing multiple common and visually-trivial relationships viz. VG-150 as compared to VrR-VG (refer Table~\ref{tab:main}). Co-localizing multiple common visual relationships requires more investigation in the line of diverse optimal solution prediction. We leave this as a future extension.
%To sum up, \prob{} is a challenging problem even for humans. We tackle \prob{} by posing it as a labeling problem and solving an optimization problem in a meta-learning framework. To verify our approach's robustness, we have provided an extensive comparison to all the relevant baselines. The results on challenging public datasets are affirmative and confirm our proposed approach's superiority over all the comparing baselines in most cases.

\section{Conclusion}
\label{sec:con}
We presented a novel task, namely a few-shot visual relationship co-localization (\prob{}), and proposed a principled optimization framework to solve this by posing an equivalent labeling problem. Our proposed model successfully co-localizes many different visual relationships with reasonably high accuracy by just looking into few images. We also show visual relationship co-localization in two more exciting settings, firstly when the subject is known in all the images, and we have to co-localize objects. Secondly, when the subject and object pair is annotated for one image in the bag, and we need to transfer this annotation to the remaining images in the bag. In both these settings, our proposed method has been found effective indicating utility of \prob{} in visual relationship discovery and automatic annotation. We firmly believe the novel task presented in this paper and benchmarks shall open-up future research avenues in visual relationship interpretation and, thereby, holistic scene understanding.

\bibliographystyle{ieee_fullname}
\bibliography{main}

\end{document}